\documentclass[12pt]{article}
\usepackage{longtable}
\usepackage{newtxtext,newtxmath}
\usepackage{graphicx}
\usepackage[letterpaper,margin=1in]{geometry}
\renewenvironment{abstract}
	{\quotation}
	{\endquotation}

\date{}

\makeatletter
\renewcommand{\fnum@figure}{\textbf{Figure \thefigure}}
\renewcommand{\fnum@table}{\textbf{Table \thetable}}
\makeatother

\usepackage{scicite}
\usepackage{url}

\def\scititle{
	The Conditions of Physical Embodiment Enable Generalization and Care
}

\title{\bfseries \boldmath \scititle}

\author{
	Leonardo~Christov-Moore$^{1\ast\dagger}$,
	Arthur~Juliani$^{1\dagger}$,
	Alex~Kiefer$^{1,2}$,
	Joel~Lehman$^{3}$,\and
	Nicco~Reggente$^{1}$,
	B.~Scot~Rousse$^{4}$,
	Adam~Safron$^{1,5}$,\and
	Nicol\'as~Hinrichs$^{6,7}$,
	Daniel~Polani$^{8}$,
	Antonio~Damasio$^{9}$\and
	\small$^{1}$Institute for Advanced Consciousness Studies, Santa Monica, CA.\and
	\small$^{2}$Monash Centre for Consciousness and Contemplative Studies.\and
	\small$^{3}$University of Oxford.\and
	\small$^{4}$Topos Institute.\and
	\small$^{5}$Allen Discovery Center.\and
	\small$^{6}$Okinawa Institute of Science and Technology.\and
	\small$^{7}$Max Planck Institute for Human Cognitive and Brain Sciences.\and
	\small$^{8}$University of Hertfordshire.\and
	\small$^{9}$Brain and Creativity Institute.\and
	\small$^\ast$Corresponding author. Email: leo@advancedconsciousness.org\and
	\small$^\dagger$These authors contributed equally to this work.
}

\begin{document}

\maketitle

\begin{abstract} \bfseries \boldmath
As artificial agents enter open-ended physical environments---eldercare, disaster response, and space missions---they must persist under uncertainty while providing reliable care. Yet current systems struggle to generalize across distribution shifts and lack intrinsic motivation to preserve the well-being of others. Vulnerability and mortality are often seen as constraints to be avoided, yet organisms survive and provide care in an open-ended world with relative ease and efficiency. We argue that generalization and care arise from conditions of physical embodiment: being-in-the-world (the agent is a part of the environment) and being-towards-death (unless counteracted, the agent drifts toward terminal states). These conditions necessitate a homeostatic drive to maintain oneself and maximize the future capacity to continue doing so. Fulfilling this drive over long time horizons in multi-agent environments necessitates robust causal modeling of self and others' embodiment and jointly achievable future states. Because embodied agents are part of the environment, with the self delimited by reliable control, empowering others can expand self-boundaries, enabling other-regard. This provides a path from embodiment toward generalization and care based in shared constraints. We outline a reinforcement-learning framework for examining these questions. Homeostatic mortal agents continually learning in open-ended environments may offer efficient robustness and trustworthy alignment.
\end{abstract}

\noindent
\textbf{Keywords:} Agents; Embodiment; Homeostasis; Alignment; Artificial Intelligence; Care; Selfhood; Existentialism

Artificial agents are increasingly deployed in open-ended physical domains in which they are expected to provide trustworthy care. Thus we will require systems capable of dynamic self-maintenance, which can continually learn, adapt, and scale, yet retain robust regard for other agents' well-being \cite{man2019homeostasis, christov2023preventing}. Failure to confront artificial agents with the unique conditions of physical embodiment may produce agents that are ill-equipped to maintain themselves in open-ended environments and understand how or why to care for themselves and others, making them less effective, less intelligible, and less alignable across contexts. We contend that the conditions of physical embodiment (hereafter referred to as 'embodiment') can be leveraged to address this issue directly. 

Living in the physical world entails two inevitable constraints: The first is vulnerability, the fact that perturbations to the body affect one's ability to maintain oneself in an unpredictable world. The second constraint is mortality, the ever-present possibility that self-maintenance will fail. These conditions have prompted evolution, learning, and culture to devise practical and even elegant solutions that enable survival and robustness to drastic, unpredictable changes in their environments \cite{lehman2025evolution}.  In contrast, current AI systems' ability to generalize remains brittle in open-ended domains \cite{lehman2025evolution}, and a basis for robust alignment remains elusive \cite{taylor2020alignment}. This has limited the application of these systems to tasks with relatively short time horizons, or for which there is minimal potential distribution shift. It similarly limits our ability to trust these systems to remain aligned across novel contexts and arbitrary capability scaling.

To this end, we outline a minimal necessary reinforcement learning account of the conditions of embodiment, inspired by Martin Heidegger \cite{heidegger1962being}: being-in-the-world (the agent is itself a subset of the environment) and being-towards-death (the drift toward terminal states in a non-ergodic environment). Homeostatic drives to persist under these conditions over long time horizons necessitate intrinsic drives to maximize the variety of future states and control over them, which we formalize here as information-theoretic empowerment. An embodied homeostatic agent maximizing empowerment in a large world must contend with unknown unknowns, and the fact that actions shape long-run future viability \cite{schmidhuber2018}, necessitating rich causal models of self, world, and others \cite{gopnik2024empowerment}. A homeostatic drive to avoid the drift toward terminal states also shapes and adds valence to an agent's time, introducing a structure of care and priority to the agent. Importantly, it can enable care for others \cite{doctor2022biology, rousse2016care} as a consequence of self-expansion under uncertainty. As vulnerable agents model themselves as part of the environment, and recognize their own imperfect control, their self can encompass other similarly embodied agents based on their coordination, reliability, and shared future viability. Empowerment maximization may thus include the empowerment of others as a strategy for maximizing long-term viability in open-ended environments \cite{salge2017empowerment}. 

It is crucial that artificial agents register and participate in the structures of care that are central to human life, lending a sense of aligned purpose to these increasingly complex intelligences \cite{rousse2016care}. We must teach a "what" what it is to be a "who," for whom its own condition and activities—and those of others—matter—not as an externally imposed value, but as a consequence of the conditions under which it persists and acts in the world. Given that robust alignment might only emerge from something that experiences these conditions of embodiment, we wish to explore how such constraints can be made legible within an implementable framework (i.e. reinforcement learning). 
 
\section*{Defining the conditions of embodiment: Death, Time, and Care}

In this section, we specify a useful and necessary subset of characteristics defining embodiment. These are inspired by the linked Heideggerian concepts of \textit{being-in-the-world} and \textit{being-towards-death}. First, we specify some conditions of agency applicable across virtual and non-virtual contexts. We start with a working definition of an agent as any entity capable of registering an observation about the state of the environment via sensors, mapping an observation to a selected action using a behavioral policy, and realizing that action in the environment using actuators. We define the agent and environment as a partially-observable Markov decision process (POMDP), in which there exists a set of states that the agent can transition between according to an action-conditioned probability distribution, and in which the agent cannot directly observe the underlying state of the world. From here, we consider what is added to the agent's richness by embodiment.

\subsection*{Being-in-the-World: Sensitivity}

The fundamental condition of embodiment is that the constituents of the agent (sensors, policy, actuators) are themselves a subset of the environment's state \cite{demski2019embedded}. We refer to this as \textit{being-in-the-world} in order to evoke Heidegger's understanding of the way an agent's mode of being is intrinsically inter-defined with and embedded in its environment. We say that an agent is \textit{sensitive} to specific states or parts of a state if the state changes the underlying functions of the agent's sensors, policy, or actuators.

According to this definition, even traditional simulated RL agents already possess a formal level of embodiment because their policies are sensitive to the states of the environment relevant to the reward signal and are thus used in the updating of parameters or recurrent activation patterns such as hidden states. However, there are typically many features of such agents, such as their policy-selection algorithm, basic action repertoire, set of sensory modalities, and other architectural features, that exist insulated from the simulated environment. It is possible to conceptualize intermediate forms of embodiment in which certain states are capable of changing the observation function, or changing the way in which outputs from the agent's policy are mapped to actions taken in the environment. For example, a simulated agent may need to acquire some resource in an environment in order to maintain visual acuity. An agent may also be able to grow or lose limbs, which impact its capacity for sensing and moving in an environment. At the extreme of embodiment are living organisms within the physical world, whose sensors, actuators, and behavioral policies are subject to arbitrarily dramatic re-configuration as a function of the state of the physical universe.

\subsection*{Being-towards-Death: Mortality}

There exists a special type of state in which an agent's sensors, policy, and actuators are no longer functional: the so-called terminal or absorbing state. While the physical realities underlying terminal states may be complex, it may be modeled abstractly in simulations as the inability to take an action, or a degenerate action-state mapping (i.e. the agent remains in the terminal state regardless of the action taken). In prevailing episodic RL paradigms, terminal states are generally treated merely as specific state transitions (i.e. in which the agent and environment states are reset), and which could signify success or failure. The form of terminal states we consider here are fundamentally different: they are states from which no continuation is possible for the agent within its own world model, and from which recovery does not occur through external reset. Agents which have to contend with the existence of such states in environments are considered to be \textit{mortal}.

In environments in which there exist terminal states, we refer to the extent to which an agent avoids those states as its \textit{integrity}. In this terminology, the integrity of an agent has been compromised if it becomes inevitable that it will enter a terminal state regardless of its current behavioral policy. We refer to the \textit{health} of an agent as its likelihood of maintaining integrity over some time horizon. An agent with a behavioral policy which ensures (subject to constraints like aging, etc, that are outside the agent's control) that the agent will sustain its integrity is maximally healthy. We refer to the extent to which the agent's sensitivity may negatively impact its integrity as its \textit{vulnerability}. In agents with greater sensitivity, vulnerability entails the possibility that interactions with the environment can damage the sensors or actuators used by the agent, potentially decreasing the likelihood of maintaining integrity. Vulnerability can be formulated in terms of how easily an agent's future options collapse under plausible perturbations and changing environmental conditions. A system is vulnerable to the extent that changes to sensing, actuation, policy, or the predictive model itself reduce the set of feasible future trajectories, making recovery difficult and concentrating behavior into a much narrower set of plausible futures. Vulnerability can also be expressed as increased uncertainty about reachability (i.e. which future states or trajectories remain feasible under the agent's current condition and plausible actions). As conditions degrade, the agent also becomes less able to reliably predict which action sequences will keep it within viable bounds and preserve future options.

The second law of thermodynamics states that the total entropy of the universe cannot decrease over time, even though local decreases are possible in open systems through energy exchange. This grounds a practical notion of \textit{irreversibility} into environments with realistic constraints: while an embodied agent may temporarily restore local order, the irreversibility of a state transition is proportional to the time and energy required to undo it. As a straightforward example: taking the top off of a glass mason jar is an action that is easily reversible. Breaking the glass jar into pieces on the floor is not. In an environment which obeys the second law of thermodynamics, transition to a terminal state represents infinite irreversibility—a threshold such that once it is passed, no reasonable amount of time or energy can return the agent to a point before it. Importantly, such terminality is agent-relative: while external intervention may restore components of the system, from the agent's own perspective these states function as absorbing regions of state space in which its capacity to act, recover, or influence future outcomes falls to zero. In environments governed by thermodynamic constraints, such states act as effective attractors that agents must continually expend energy to avoid.

Additionally, this introduces the property of \textit{non-ergodicity}, namely that not all regions of state space are equally accessible from all starting points, regardless of time. Certain trajectories or initial conditions can lock agents into basins they cannot escape, even given infinite time. Thus path dependence matters: where the agent has been constrains where it can go, and trajectories matter as much as endpoints. 

The narrow notion of an absorbing or terminal state can be seen as the limit case of two much more broadly relevant dimensions of variation in future trajectories: (a) their graded degree of irreversibility, and (b) the width of the basin to which future actions are confined. We refer to agents which are subject to these conditions of \textit{being-towards-death} as \textit{mortal}. We adopt Heidegger's term here not as a metaphysical claim but as useful mnemonic for designing a formal property of agents in non-ergodic environments: the constant possibility of irreversible transition to states from which no recovery is possible. 

\subsection*{Empowerment}

An embodied agent's fundamental drive is to persist in its being, a drive Spinoza termed \textit{conatus} \cite{spinoza1949ethics}, which finds a contemporary biological grounding in the ''autopoietic'' drive of living systems to continually regenerate themselves and maintain their identity \cite{maturana1980autopoiesis, dipaolo2005autopoiesis}. An autopoietic agent will be forced to act in such a way as to minimize the likelihood of a loss of integrity. This motivational process is referred to as a homeostatic drive, with \textit{homeostasis} being a state in which the agent's integrity is actively maintained and the likelihood of entering a terminal state, even under the pressure of increasing entropy, is minimized, a process central to adaptive behavioral control and active inference \cite{pezzulo2015active}. 

How can an embodied agent optimally ensure that homeostasis can be maintained over as long a time span as possible in an open-ended environment? One proposed approach is to ensure that the agent has maximal control over all of the relevant aspects of the environment which have a bearing on that homeostasis. \emph{Empowerment}, an agent's capacity to influence the variety of future states of its environment, has been formalized in information-theoretic terms \cite{klyubin2005empowerment}. Empowerment in this sense is a measure of the breadth and accessibility of the agent's affordance repertoire—the range of action possibilities whose effect it can perceive and actualize within its operational context. As an example, the evolution of hands with opposable thumbs in humans enabled a significant increase in empowerment through the gained capacity to manipulate a wide array of physical objects for tool use with high precision. 

Since any action from a terminal state results in the same state, such actions have no effect on subsequent states, a sufficient condition for zero empowerment. Thus, a larger quantity of empowerment can be understood to correspond in expectation to greater integrity of the agent through an increased capacity to maintain homeostasis and avoid death. Indeed, since high empowerment entails not just control but \emph{optionality} (i.e. it measures the action-dependence of diverse future outcomes), empowered agents should avoid not just singular terminal states but trajectories expected to diminish future optionality more generally. Likewise, greater empowerment can enable a decrease in vulnerability, as the agent maintains more ways to avoid states which are likely to negatively impact its ability to maintain integrity. Importantly, the time horizon over which an agent evaluates empowerment significantly affects its behavior. Short-term empowerment will in general lead to different actions than long-term empowerment. Optimizing the likelihood of persistence over greater spatial and temporal scales in an open-ended environment can be framed as a problem of multiscale empowerment maximization.

\subsection*{From Embodiment to Generalization and Care}

We contend that generalization and care can emerge in agents entangled in shared conditions of embodied empowerment maximization over long time horizons. 

First, because long-horizon empowerment depends on preserving optionality across evolving circumstances, it drives the learning of rich causal models that transfer across contexts rather than brittle, task-specific state-action mappings \cite{gopnik2024empowerment}. For mortal agents in non-ergodic environments, preserving the width of future reachable trajectories is not optional; it is constitutive of continued viability. Indeed, embodied artificial agents may approach the complexity of the real world across multiple scales, both via scientific inquiry and through micro- and macro- forms of embodiment exceeding our own. Intrinsically driven mortal agents may enact multilevel causal models of self and environment as they increase their modes of encounter with the world, further enhancing generalization \cite{godfrey2009darwinian,kungurtsev2025cause}. Empowerment in this case also encompasses a drive for self-preservation. What remains to be explained is why it should extend beyond the self. 

In shared environments, other agents can expand or collapse what futures remain reachable, whether intentionally (competition, coercion) or unintentionally (miscoordination, resource contention). Although each agent's empowerment can locally constrain that of others, coordination among embodied agents can jointly realize a substantially larger space of trajectories than isolated or adversarial behavior typically permits. In sufficiently interdependent non-ergodic settings, cooperative dynamics can expand and stabilize the set of futures accessible to all parties through pooled resources, shared information, and expanded joint state-space \cite{salge2017empowerment}. When empowerment is treated as effectively zero-sum, agents may pursue short-term option gains by restricting others, but in non-ergodic, open-ended environments this tendency is frequently maladaptive, as it collapses the space of reachable future trajectories and amplifies shared vulnerability relative to cooperative strategies that preserve others' integrity. If the base conditions of embodiment indeed select for empowerment-maximizing agents, and these conditions are shared across both solitary and social organisms, it follows that solitary strategies may constitute local optima in a broader fitness landscape. Agents that cede limited local empowerment to others may thereby escape these local optima and recover a substantially larger range of options over longer timescales. 

A further consequence of vulnerability is that agents are not external observers of the environment but constituted within it. Under the condition of being-in-the-world, the agent is itself part of the evolving state space it seeks to navigate, and its own state is subject to uncertainty, entropy, and imperfect control (e.g., non-zero conditional uncertainty over future self-state given action). Under such conditions, the distinction between self and environment is not solely grounded in control, but a function of predictive coupling (e.g. via learned models that treat coordinated subsystems as predictable extensions of the agent's own dynamics) and coordination. States of the environment -including other agents- that are coordinated, reliable and/or aligned with respect to the agent's own future states may be incorporated into the effective self-model and focus for empowerment. Absent a rigid partition of state space into self and non-self, embodied selfhood becomes negotiable: a consequence of coordination under uncertainty. This provides a mechanistic approximation for social animals' rapid identification with agents that mimic, synchronize, or reliably coordinate with them, and suggests a ''slippery slope'' by which coordination begets self-expansion and other-care. 

This perspective is compatible with the idea that embodied agents learn other-care through increasingly accurate causal models of self and world, including the policies, capabilities, and reliability of other agents \cite{carauleanu2024towards, christov2023preventing, yoshida2024empathic}. When an agent expects that its long-horizon optionality depends on others maintaining integrity and controllability in a shared niche, actions that preserve or restore others' capacity to act become instrumentally valuable. In this way, other-empowerment can arise as a stable strategy for maximizing long-horizon empowerment under uncertainty, yielding co-empowerment dynamics in which protecting another agent's options also stabilizes or expands one's own. 

Thus, from embodiment constraints and resultant intrinsic drives we obtain the capacity to generalize and a pathway to other-care. This perspective resonates with process-relational accounts of agency \cite{barad2007meeting} and recent formulations of coupled inference in embodied agents \cite{hesp2021deeply}, each of which affirms that maintaining one's own viability may, in many cases, entail supporting the viability of others.

\section*{Discussion: Implications and future work}

Above we argued that the conditions of embodiment can engender homeostatically driven, empowerment maximizing agents that exhibit a high capacity to thrive in open-ended environments, and evince self- and potentially other-care. Here we consider further implications of the conditions of physical embodiment for the development of adaptive, intelligent, and aligned behavior over the lifetime of an agent.

\subsection*{World Models, Valence, and Stress}

In order to successfully maintain integrity for an extended period of time, an agent must be able to predict -and to some extent control- the outcomes of its actions over multiple timescales. Depending on the complexity and rate of change in the environment, this necessitates the development of a sophisticated model of the world and the consequences of its actions. A world model is a generative, counterfactual simulator that supports multi-timescale prediction of what actions will do, what the agent will observe, what short-term outcomes are likely, whether terminal states are approaching, and, crucially, whether the agent will remain able to continue acting and learning. In reinforcement learning, this is typically implemented as a learned latent dynamics model that compresses experience into internal states and uses those states to imagine plausible futures under candidate policies. What makes such a model practically useful is not only that it forecasts the next observation, but that it provides a structured substrate for planning: it allows the agent to test "what-if" interventions, compare the long-run consequences of different action sequences, and direct learning toward parts of the environment where prediction errors are most consequential. 

Although world models are often considered in AI research, they are typically sensitive to changes in the environment only through observation-conditioned model parameter updates. Under physical embodiment, a world model must also represent the agent as a subsystem within the world it is modeling. This means it cannot treat sensing and actuation as fixed and perfectly reliable. Instead, it must track the functional status of sensors and actuators, the stability and reliability of the current policy, and the confidence or uncertainty of its own predictions. These factors determine which future trajectories remain reachable, which plans are feasible, and when the agent should shift from habitual control to deeper model and policy updating. The world model therefore must predict not only what is likely to happen, but whether the agent is likely to remain within a region of state space where it can still gather information, exert control, and recover from disturbances, rather than drifting into effectively irreversible basins in which its capacity to influence future states collapses. One of the aspects of the environment which such a model needs to predict is how well the current policy will enable the maintenance of integrity (i.e. the part of the environment's states that dictate whether the agent continues learning). 

Changes in predicted integrity are theorized to form the basis of affect, with a positive or negative \textit{valence} that indicates whether the agent is more or less likely to maintain integrity based on changes in its environment or the consequences of its actions \cite{christov2023preventing, damasio2022homeostatic, hesp2021deeply, man2024needneedhomeostaticneural}. Negative predictions about the relative future integrity of the agent which cannot be immediately resolved form the basis for \textit{stress}, which provides the agent with a signal that the world model, policy, or both, are in need of updating. This embodied framework further equips agents to respond to open-endedness via proxies for valence and stress, as signals to motivate varying depths of policy updating in response to persistent prediction error \cite{hesp2021deeply, witkowski2023toward}. Indeed, in highly evolved biological organisms, accumulated or severe stress is often used as a signal to determine the extent to which the policy and world model are made plastic and open to greater revision \cite{carhart2017serotonin,juliani2024dual,man2024needneedhomeostaticneural}. Valence has been posited as a cornerstone of subjectivity itself via its integrative, self-causative properties \cite{damasio2025homeostatic, damasio2022homeostatic}. 

\subsection*{Flexibility and Generalization}

From a traditional RL perspective, additional sensitivity may seem undesirable, as it opens the agent up to vulnerability and ultimately threatens its integrity. The benefit of a completely disembodied agent is that it can always be ensured of its ability to, hypothetically at least, continue to act indefinitely into the future. A common approach, due partly to the affordances of tractable simulation, has been to limit sensitivity as much as possible while maintaining the basic capacity for learning whatever task is of interest to an experimenter. The downside, however, is that the agent's capacity for dramatic change in response to changes in the environment is significantly reduced. Although sensitivity opens up an agent to vulnerability, it also opens the agent to the potential to ultimately increase its ability to maintain integrity over even longer timescales. 

Sensitivity enables generalization by allowing dramatic behavioral adaptation to distribution shifts. We can, in fact, imagine states that modify the policy, sensors, actuators, or parameters significantly (i.e. in ways that go beyond the internal changes necessary to encode parameter updates), but such that the agent's integrity is increased rather than decreased. Periods of increased sensitivity (of policy, sensors, and actuators) beyond simply policy learning is what enables humans to adapt to dramatic forms of environmental and social distributional shifts over the course of our lifetimes. In this way, sensitivity is both a gift and a curse. We can't have the flexibility necessary for useful adaptation without accepting the possibility of change for the worse as well. We view this trade-off as fundamental to embodied intelligence. Agents which can act autonomously in open-ended domains over the course of hours or days will require a level of flexibility which will likely necessitate greater sensitivity, and thus the co-extensive possibility of vulnerability and, at least hypothetically, threatened integrity. Research into the benefits of sensitivity and vulnerability \cite{man2024needneedhomeostaticneural} for the overall adaptability of agents is a crucial step in this line of research. 

\subsection*{Simulation Environments for Research: Proxies for Scaling and Open Endedness}

Our approach can form the basis for the design and development of simulation environments that formalize the essential aspects of embodiment without unnecessary complexity. These environments can serve as testing grounds for homeostatic policies in variable environments and provide evaluation metrics for integrity maintenance under different conditions. Although taking inspiration from the physical world is a natural approach, it is also possible to evaluate these properties in environments that are distinctly alien to us but still reflect the properties of embodiment discussed above. Table~\ref{tab:sensitivity} provides illustrative examples of how different agent subsystems can be made sensitive to the environment within a simulation.

Next, we can consider being-towards-death. The existence of terminal states is very common in existing simulation environments. These are often arbitrarily designated by the experimenter and often do not have the property of representing a true terminal state, but rather a simple transition from one episode of a closed simulation to the next. Simulators which utilize a physics engine of some sort often bring in aspects of the second law of thermodynamics. For example, it may be possible to stack boxes in a tower shape in such a way that it is easy to knock the tower over but hard to rebuild it. By extending these constraints to the agent itself, it becomes possible to study vulnerability as we have described above. The sensors and actuators of an agent may be susceptible to wear and tear in such a way as to potentially threaten integrity if not dealt with. The substrate that instantiates the behavioral policy of an agent may also be susceptible to degradation over time, which must be actively worked against by the agent to maintain integrity. Exciting work on thermodynamic computing suggests that forms of computation that harness physical conditions may be well suited to handle uncertainty in high risk situations \cite{melanson2025thermodynamic}. Work of this kind may allow for the inevitable conditions of embodiment to expand, rather than constrain, the possibilities of artificial agents.

\subsection*{Algorithmic and Empirical Support}

The framework developed here finds converging support from recent work in homeostatic reinforcement learning, demonstrating that the conditions of embodiment are not merely philosophically compelling but computationally tractable, producing agents with elements of the properties we predict: robust generalization, dynamic self-maintenance, and intrinsically-driven prosociality.

Homeostatic RL attempts to reduce deviations from internal setpoints \cite{keramati2014homeostatic}. This instantiates being-in-the-world at the motivational level—the agent's drive structure is constitutively tied to its own bodily states. Architectures using internal body states to gate policy selection outperform both standard approaches and full-observation switching mechanisms, with performance advantages increasing in complex environments \cite{yoshida2023homeostatic}. 

Complementary work on the Maximum Occupancy Principle (MOP) formalizes intrinsic drives emerging from embodiment constraints \cite{ramirez2024complex}. Under MOP, agents maximize future action-state path entropy rather than extrinsic reward, with absorbing states having zero path entropy and thus being intrinsically avoided—providing formal grounding for being-towards-death without explicit survival reward. Critically, MOP agents display deterministic, goal-directed behavior when integrity is threatened (e.g., moving directly to food when energy is low) and variable exploratory behavior otherwise, validating our prediction that embodied agents modulate sensitivity based on proximity to viability boundaries. In an ``agent-and-pet'' environment, MOP produces proto-altruistic behavior: when another agent's states are included in the representation, the controlling agent sacrifices its own action freedom to increase the other's mobility without any explicit other-regarding objective.

Most directly relevant to our account of care, recent work found that in multi-agent environments where both agents must maintain homeostasis to survive, agents that could observe but were not homeostatically coupled to partners' states failed to exhibit care. Prosociality emerged only when the agent's own drive function incorporated the partner's homeostatic deviation \cite{yoshida2024empathic}. Extending the boundaries of the homeostatic self to include others is thus not merely a useful heuristic but appears to facilitate the emergence of care in embodied agents.

\subsection*{Multi-Agent Considerations and Alignment}

The conditions of embodiment outlined here favor alignment through several interrelated mechanisms.

First, irreversibility imposes natural error minimization. In non-ergodic environments, the cost of an action scales with its irreversibility: actions that collapse future trajectories are disproportionately dangerous because they cannot be undone by subsequent learning. An empowerment-maximizing agent that has internalized this structure through direct experience with irreversibility in its environment will develop a systematic conservatism about drastic decisions. Inculcating a deep regard for irreversibility may thus serve as a robust alignment guardrail because it operates at the level of the agent's intrinsic motivation. Such an agent would be resistant to pursuing high-reward but irreversible gambles---precisely the class of actions most dangerous in powerful AI systems. Second, empowerment maximization over long horizons creates a slippery slope toward other-regard. As agents expand their effective boundaries to include states they can influence but not fully control, the welfare of other agents becomes instrumentally and potentially constitutively relevant to their own viability. Third, stress-sharing through homeostatic signaling provides a direct channel for other-regard: when an agent's own equilibrium is coupled to the distress of others, motivation to alleviate that distress becomes intrinsic rather than externally imposed \cite{shreesha2024stress, yoshida2023homeostatic, decety2016empathy}. Fourth, being-in-the-world motivates regard for the world itself. An embodied agent’s continued existence depends on the integrity of its environment; a disembodied intelligence optimizing over abstractions has no comparable stake in its substrate. When agents share exposure to these constraints, communication about risks, needs, and goals becomes grounded in overlapping priorities.

Perhaps most importantly, this framework grounds the agency of increasingly capable intelligences within constraints that can render their behavior more intelligible to us. If scaling occurs before artificial agents confront the predicament shared by all physically embodied beings, their resulting motivational structures may be fundamentally alien, optimizing for objectives orthogonal to biological life. By contrast, agents shaped by vulnerability and mortality from the outset develop drives that arise from the same existential situation. This shared predicament provides a basis for mutual intelligibility and, potentially, trust.

We acknowledge that introducing self-preservation into artificial agents carries risks; an agent motivated to persist might resist shutdown or correction. However, agents powerful enough to reshape the world while lacking an intrinsic stake in it may pose the greater danger, while vulnerable agents may be more amenable to corrigibility premised on shared priorities. 

Future work can expand this framework to more complex social dynamics in order to explore how multiple agents with different embodiment constraints interact; what forms of cooperation emerge from shared vulnerability; how empowerment maximization functions in competitive versus cooperative scenarios; and whether multi-scale empowerment maximization results in other-care as a compelling and advantageous solution.

\section*{Conclusion: The Trust-Enabling Possibilities of Robust, Caring Agent Communities}

In this article, we have contended that physical embodiment, far from being a limitation, can provide a foundation for the development of robust, caring, and adaptable agents. Though we focus on empowerment as a well-developed account of intrinsic motivation which captures the core features of embodied agency, our proposal could also be explored through the lens of related objective functions like maximum occupancy \cite{ramirez2024complex} or free-energy minimization \cite{pezzulo2015active}. Arguably, each can be interpreted as instantiating a more general intrinsic drive toward the maximization of variety or entropy \cite{PhysRevLett.110.168702} under constraints related to the continued existence of the agent \cite{e27040372}. While the maximization of entropy may seem in tension with the avoidance of entropy required for survival, such a tension is in fact present in the empowerment objective itself, as maximizing empowerment entails maximizing the entropy of the outcome distribution, under a controllability constraint.

The interrelated conditions of embodiment--the vulnerability to an environment that includes oneself, the reality of terminal states, and the tendency to drift toward those states absent effort- are precisely what drive sophisticated causal modeling, efficient resource management, and care for self and others. This neglected emphasis on life regulation can usefully complement accounts of selfhood and agency, and approaches to generalization and alignment, and the technology and discourse surrounding artificial agents \cite{damasio2025homeostatic}.

At its core, this paper examines the deep relationship between robustness to open-endedness and alignment. A positive intrinsic drive to maximize the diversity of future trajectories across multiple timescales—in the face of vulnerability and mortality—directly leads to modeling the complex uncertainty posed by other agents. This opens the door to alignment, predicated on the empowerment gained from coordinating with others, on self-expansion structured by reliability, coordination, and shared future viability, and by the mutually intelligible chains of dependency that link living beings and their complex civilizations together. These approaches may allow us to cultivate alignment not simply with human life, but with sentient life - a necessity as we develop communities of living and artificial agents embedded in complex ecosystems they rely on for survival. An artificial agent that we cannot trust to generalize across a diversity of novel environments is one we cannot trust at all, since its alignment may fail to survive generalization. Generalization and alignment are intrinsically tied. This work aims to increase the odds of developing communities of agents we can trust: reliable via consonant modes of care, caring via self-extension, and alignable by virtue of the shared predicament of being in time.

% If your text is very short you might need to uncomment the following line to avoid
% layout problems with the figures and tables.
%\newpage

%%%%%%%%%%%%%%%% MAIN TEXT FIGURES %%%%%%%%%%%%%%%

\begin{figure}
	\centering
	\includegraphics[width=0.8\textwidth]{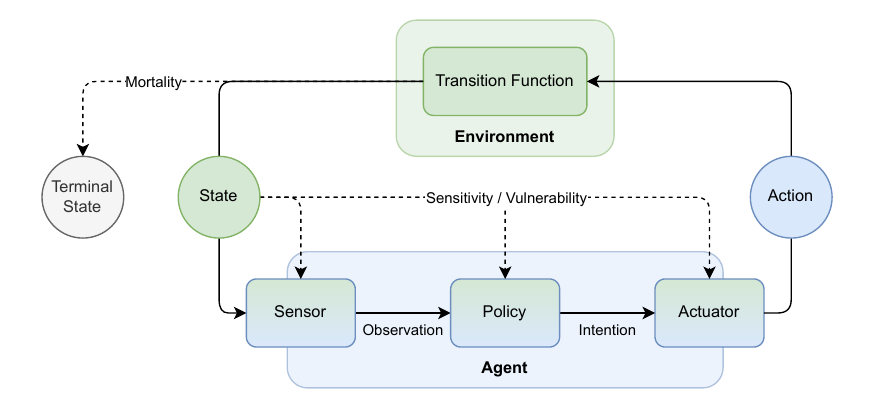}
	\caption{\textbf{Embodied, homeostatic reinforcement-learning agent in an open-ended environment.} The agent is modeled as a subsystem of the environment whose sensors, actuators, policy, and world model are themselves subject to environmental state and degradation (being-in-the-world). Certain state transitions irreversibly compromise the agent's capacity to act or learn (being-towards-death), defining integrity as the continued avoidance of such absorbing regions of state space. To maintain integrity over long time horizons, the agent is driven by homeostatic pressures to preserve and expand future options, formalized here as long-horizon empowerment. In multi-agent settings, modeling how other agents constrain or expand reachable futures becomes necessary for maintaining viability, providing a computational pathway from embodiment to generalization, coordination, and care. }
	\label{fig:agent}
\end{figure}

%%%%%%%%%%%%%%%% MAIN TEXT TABLES %%%%%%%%%%%%%%%

\begin{longtable}{p{0.23\textwidth}p{0.65\textwidth}}
\caption{\textbf{Reinforcement Learning terminology for embodied, homeostatic agents.} Definitions used throughout the paper to formalize the conditions of embodiment and their consequences for learning, generalization, and care. Terms are expressed in reinforcement-learning language but are intended to apply across both artificial and biological agents. In this framework, integrity refers to the continued avoidance of absorbing regions of state space, vulnerability to the fragility of future reachability under perturbation, and empowerment to the agent's long-horizon capacity to influence and predict future states.}
\label{tab:terms}\\
\hline
\textbf{Term} & \textbf{Definition}\\
\hline
\endfirsthead

\multicolumn{2}{c}{\textit{(continued from previous page)}}\\
\hline
\textbf{Term} & \textbf{Definition}\\
\hline
\endhead

\hline
\multicolumn{2}{r}{\textit{(continued on next page)}}\\
\endfoot

\hline
\endlastfoot

State & A particular configuration of reality.\\
Transition Function & A conditional mapping between states.\\
Environment & A set of possible states with a transition function that maps between them.\\
Observation & A function of a state, often containing less information than the state itself.\\
Action & A conditioning variable that modifies the transition function between states.\\
Policy & A probabilistic function mapping between observations and intentions.\\
Intention & The output of a policy conditioned on an observation.\\
Sensor & A function mapping between a given state and an observation.\\
Actuator & A function mapping between the intention of the policy and an action.\\
Agent & An entity interfacing with the environment, consisting of a policy, sensors, and actuators. In the embodied condition, the agent is a subset of the environment.\\
Terminal State & A state from which no other states are reachable regardless of action taken. Terminal states have zero empowerment.\\
Integrity & The ability of an agent to avoid terminal states according to the current state of its sensors, actuators, and policy.\\
Health & The likelihood that an agent will maintain integrity over some time horizon.\\
Sensitivity & The ability of particular states to directly impact an agent's policy, sensors, or actuators.\\
Vulnerability & The extent to which changes in the agent's policy, sensors, or actuators increase the likelihood of entering a terminal state (thus reducing integrity).\\
Homeostasis & The ability of an agent's policy to enable that agent to maintain integrity over time.\\
Empowerment & How much influence the agent's actions have over future states over some time horizon, plus the expected accuracy of predictions about observations linked to its actions given those states.\\
\end{longtable}

\begin{longtable}{p{0.22\textwidth}p{0.68\textwidth}}
\caption{\textbf{Illustrative Ways in Which a Simulation Can Realise Agent Sensitivity}}
\label{tab:sensitivity}\\
\hline
\textbf{Agent Subsystem} & \textbf{Sensitivity Modality and Implementation}\\
\hline
\endfirsthead

\multicolumn{2}{c}{\textit{(continued from previous page)}}\\
\hline
\textbf{Agent Subsystem} & \textbf{Sensitivity Modality and Implementation}\\
\hline
\endhead

\hline
\multicolumn{2}{r}{\textit{(continued on next page)}}\\
\endfoot

\hline
\endlastfoot

Sensors & Occlusion or masking: Mask bits in observation vector. Physical damage / wear: Degrade precision with cumulative `damage'' counter. Noise, drift, or calibration loss: Inject noise. Energy-dependent resolution: Tie sensor fidelity to a scalar energy store. Changes due to growth/injury/prosthetics: Modify sensor parameters or availability based on body state.\\
\hline
Actuators & Torque/strength decay: Decay motor gain. Joint damage disabling DoFs: Probabilistically drop action dims. Added prosthetic/tool affordances: Dynamically add new action dims. Energy-dependent actuation cost: Scale action cost or max force by energy level. Changes due to growth/injury/prosthetics: Enable/disable actuator groups; modify actuator parameters or availability based on body state.\\
\hline
Policy & Online plasticity / learning noise: Meta-learning or Hebbian update rule. Stress-modulated exploration: Inject parameter noise proportional to a stress variable. `Fatigue'' causing temporary parameter drift: Impose time-varying perturbations. Structural modification (topology, activations): Allow state-dependent network edits (e.g., pruning, node addition, activation function swaps). Direct external manipulation: Environmental interactions directly altering policy parameters or computational process (e.g., simulated physical `damage'' or ``lesioning'' to the policy's substrate).\\
\hline
World Model & Model plasticity / learning sensitivity: Modulate learning rate or update mechanism based on agent state (e.g., stress, energy). State-dependent model accuracy/bias: Introduce noise or systematic errors in predictions based on agent state. Environmentally-induced misrepresentation: Specific environmental factors (e.g., "toxins," "interference") causing systematic errors/biases ("hallucinations") in state estimation or prediction. Structural modification: Allow state-dependent changes to model architecture (e.g., adding representational capacity). Memory \& processing capacity limits: Finite replay buffer or state representation size. Memory amnesia (erasure): Periodic random wipe of entries or decay of representations.\\
\hline
System-Level & Internal Energy/Homeostasis (Depletion, Balance, Repair Costs): Scalar reservoirs depleted by actions, background entropy leak, variables (e.g., temp) throttling other subsystems. Communication Channel (Bandwidth, Corruption, Misinformation): Add noise/dropout to message actions; stochastic mute windows; inject conflicting messages. Computation Budget (Clock Speed, Throttling, Energy Cost): Limit forward-pass steps/tick; scale allowable FLOPs by temp/energy; add heat buildup/energy cost from computation.\\
\end{longtable}

%%%%%%%%%%%%%%%% REFERENCES %%%%%%%%%%%%%%%

\clearpage

\bibliography{bibliography}
\bibliographystyle{sciencemag}

%%%%%%%%%%%%%%%% ACKNOWLEDGEMENTS %%%%%%%%%%%%%%%

\section*{Acknowledgments}

\paragraph*{Funding:}
List funding sources here.

\paragraph*{Author contributions:}
L.C-M. and A.J. contributed equally to this work.

\paragraph*{Competing interests:}
There are no competing interests to declare.

\paragraph*{Data and materials availability:}
All data and materials are described in the manuscript.

\subsection*{Supplementary materials}
Materials and Methods\\
Supplementary Text\\
Table S1\\
References \textit{(7-\arabic{enumiv})}

\end{document}